\documentclass[letterpaper, 10pt, conference]{ieeeconf}

\IEEEoverridecommandlockouts                        
\usepackage[hyphens]{url}
\usepackage{graphicx}
\usepackage{cite}
\usepackage{booktabs}
\usepackage{multirow}
\usepackage{amsmath,amssymb,mathtools}
\usepackage{amsthm}
\usepackage{algorithm}
\usepackage{algorithmic}
\usepackage{balance}
\usepackage{stfloats}
\usepackage{placeins}

\newtheorem{theorem}{Theorem}
\newtheorem{proposition}{Proposition}
\newtheorem{assumption}{Assumption}

\newcommand{\tabfont}{\footnotesize}
\renewcommand{\arraystretch}{1.12}
\newcommand{\tabnote}[2][\linewidth]{\par\vspace{2.5pt}\parbox{#1}{\tabfont #2}}

\title{Anticipatory Robot Goalkeeping via Monotone Optimal Stopping}

\author{Hao E. Zhang$^{*,1,2}$, Ruize Geng$^{*,1}$, Yisen Li$^{2}$, Yaru Niu$^{1}$, Yikai Wang$^{1}$, Raihan Haque$^{3}$, Khalil Zbiss$^{3}$,\\ Guanyang Luo$^{3}$, Hui-ping Wang$^{3}$, H. Eric Tseng$^{\dagger,2}$ and Ding Zhao$^{\dagger,1}$
\thanks{$^{*}$Equal contribution.}
\thanks{$^{\dagger}$Correspondance to Ding Zhao (dingzhao@cmu.edu) and H. Eric Tseng (hongtei.tseng@uta.edu).}
\thanks{$^{1}$Hao E. Zhang, Ruize Geng, Yaru Niu, Yikai Wang, and Ding Zhao are with Carnegie Mellon University.
        (email: haoz4@andrew.cmu.edu; rgeng3@jh.edu; dingzhao@cmu.edu)}%
\thanks{$^{2}$Hao E. Zhang, Yisen Li and H. Eric Tseng are with the University of Texas at Arlington.
        (email: haoz4@andrew.cmu.edu; hongtei.tseng@uta.edu)}%
\thanks{$^{3}$Raihan Haque, Khalil Zbiss, Guanyang Luo and Hui-ping Wang are with General Motors.
        (email: raihan.haque@gm.com; khalil.zbiss@gm.com; guanyang.luo@gm.com; hui-ping.wang@gm.com)}%
}

\begin{document}
\maketitle
\thispagestyle{empty}
\pagestyle{empty}

\begin{abstract}
Robots engaged in fast physical interactions often need to act before the intent of another agent is fully known. Anticipatory goalkeeping illustrates this challenge. Waiting provides more reliable information about the target but reduces the physical opportunity for interception, whereas acting early preserves reachability but requires initiating motion under uncertainty. Given a fixed closed-loop save controller, we formulate the decision of when to initiate motion as a policy-conditional finite-horizon optimal stopping problem. Building on this formulation, we propose monotone optimal stopping (MOS), a structured release-timing method for dynamic robotic interception. The quadruped save policy is trained with reinforcement learning, while MOS determines when the policy should be activated from the evolving robot state and target belief. Rather than predicting a release time or relying on confidence alone, MOS learns the return advantage of acting now over waiting for one more observation. We derive a direct Bellman recursion for this act-versus-wait margin and impose monotonicity only with respect to physical urgency, reflecting the irreversible loss of interception opportunity as time elapses. This structure enables early activation for dynamically demanding saves while preserving closed-loop adaptation when later observations change the predicted target. Under a single-crossing condition, MOS admits a threshold release boundary with a bounded approximation error. Extensive simulation studies show that MOS improves the mean save rate from $67.7\%$ to $74.4\%$ over a parameter-matched learned gate and increases reversal saves from $52.1\%$ to $66.5\%$. Real-robot experiments further demonstrate rapid interception and post-release direction correction under human shot-direction feints.
\end{abstract}

\section{Introduction}

\begin{figure}[t]
    \centering
    \includegraphics[width=\linewidth]{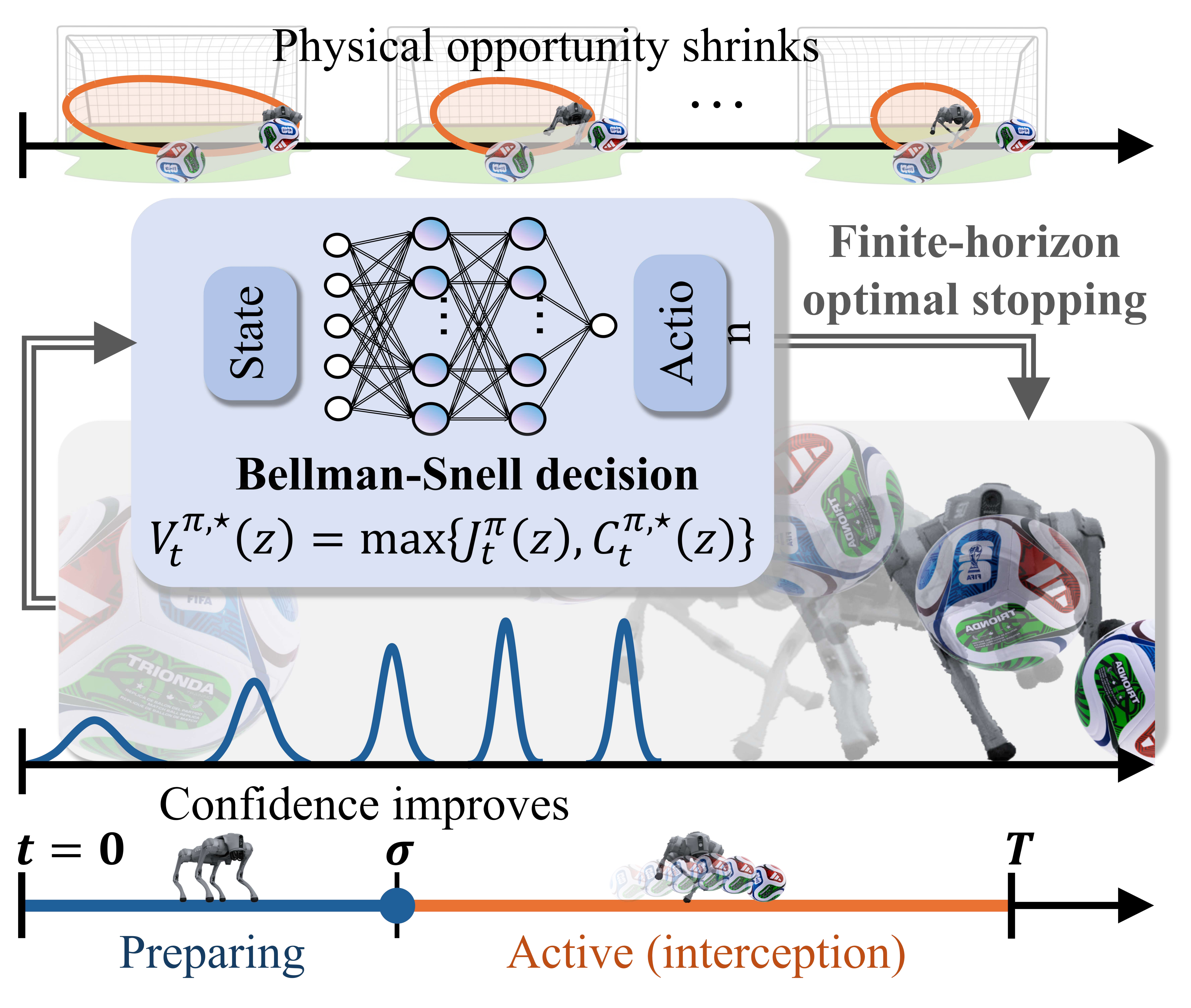}
    \caption{Anticipatory goalkeeping is posed as an optimal-stopping problem in which target information improves while the remaining motion opportunity shrinks before contact; at each decision, the robot either activates the closed-loop save policy with $J_t^\pi$ or waits one interval with continuation value $C_t^{\pi,\star}$, and release occurs when acting becomes more valuable than waiting while the save policy remains closed-loop after activation.}
    \label{fig:overview}
\end{figure}

A robot goalkeeper must answer two questions: where should it move, and when should it start moving? The second question is difficult because information and physical opportunity change in opposite directions. Before the kick, the target is uncertain but the robot has the most time to accelerate and cover the goal. Waiting provides better evidence about the shot, but it also consumes the motion time needed to reach distant targets \cite{wang2017anticipatory,zhang2026interaction}. If the robot waits until the target is certain, a corner save may already be physically impossible. Anticipatory goalkeeping is therefore an act-before-certainty problem: the robot must decide when the value of acting early exceeds the value of one more observation without losing the ability to redirect if later cues, including human feints, overturn the initial target. Fig.~\ref{fig:overview} summarizes this information--opportunity trade-off and the one-way release decision studied in this paper.

Existing quadruped goalkeepers compose skill-specific controllers under high-level planning \cite{zhang2026cognition}, while recent humanoid goalkeeping learns unified reactive control from task-motion constraints \cite{ren2025humanoidgoalkeeper}. Agile learned soccer control has also advanced rapidly \cite{haarnoja2024soccer}. These systems mainly focus on how to execute a save once informative ball motion is available. Related early-prediction methods determine when sufficient evidence has been acquired \cite{chen2020stop}, but confidence alone does not capture the motion opportunity lost by waiting \cite{zhang2025multi}. A fixed trigger ignores whether the current robot state can still reach the target, while an unconstrained learned gate has no reason for its act/wait boundary to follow the one-way loss of physical opportunity \cite{huang2023goalkeeper}. The relevant question is therefore not simply whether the shot prediction is confident enough, but whether waiting one more step is still worth the motion opportunity that will be lost.

Our key design choice is to separate when to act from how to save. We first train a closed-loop save policy with proximal policy optimization (PPO) to map recurrent robot and pre-contact observations to joint-position commands under interception and stability rewards. After training, the motor policy is frozen and MOS learns only its release time. Before release, a ready controller keeps the robot recoverable; after release, the save policy continues to observe the evolving target belief and may redirect if the early cue changes. Thus early activation preserves motion time without fixing the target, allowing closed-loop correction under cue reversals and human feints.

We formulate this release decision as finite-horizon optimal stopping \cite{peskir2006optimal} and propose monotone optimal stopping (MOS). MOS learns a stopping margin equal to the return of acting now minus the return of waiting one more decision interval. The Bellman--Snell recursion gives a direct target for this margin, so the learned quantity is exactly the quantity whose sign determines release. We then impose one physical structure: for the same robot state, target belief, and decision time, increasing urgency should not make waiting more attractive again. The resulting critic is monotone only along urgency and remains unrestricted in all other inputs. Consequently, MOS can release earlier for dynamically difficult shots and later for easy shots while avoiding irregular wait--act--wait boundaries as the interception window closes.

The paper makes three contributions: 1) a policy-conditional finite-horizon optimal-stopping formulation that converts anticipatory release into an explicit comparison between acting now and waiting for one more observation, together with a direct Bellman recursion for the resulting act-versus-wait margin; 2) monotone optimal stopping learns this margin without binary timing labels and imposes monotonicity only with respect to physical urgency, yielding a threshold-structured release boundary under a strong single-crossing condition together with an approximation-to-boundary error bound; 3) a MOS-guided quadruped goalkeeping system that integrates anticipatory timing with a PPO-trained closed-loop save policy, improving mean save rate from $67.7\%$ to $74.4\%$ and matched reversal save rate from $52.1\%$ to $66.5\%$ over a parameter-matched learned gate, while real-robot shot-direction feints demonstrate post-release correction after misleading early cues.

\section{Related Work}

\subsection{Legged interception and goalkeeping}
Large-scale reinforcement learning (RL) has enabled agile locomotion and ball interaction on legged robots \cite{rudin2022walk,haarnoja2024soccer}. The quadrupedal goalkeeper in \cite{huang2023goalkeeper} composes skill-specific controllers under a high-level planner, while recent humanoid goalkeeping learns a unified reactive controller with region-conditioned motion priors \cite{ren2025humanoidgoalkeeper}. These systems demonstrate increasingly capable interception once task-relevant ball motion is available. Our focus is the preceding decision: whether a prepared robot should remain in a recoverable ready mode or activate a save while the future target is still represented by a belief.

\subsection{Anticipatory control under partial information}
Sequential decision making under incomplete observations is classically modeled through belief-state control \cite{kaelbling1998pomdp,lauri2023pomdp}. In fast physical tasks, however, information acquisition and control authority evolve simultaneously. Anticipatory table tennis explicitly trades observation against preparation time \cite{wang2017anticipatory,zhang2026halo}, and early-prediction methods learn when sufficient evidence has been acquired \cite{chen2020stop}. Our setting differs in that waiting changes the feasible physical response of a high-dimensional controller; confidence alone therefore does not represent the value of another observation.

\subsection{Optimal stopping and monotone function approximation}
Finite-horizon stopping is characterized by the Snell envelope and the boundary between continuation and stopping regions \cite{peskir2006optimal,zhang2025bi}. Approximate dynamic programming and deep stopping methods estimate values or policies from sampled trajectories \cite{tsitsiklis2001regression,becker2019deep}. We instead learn the act-versus-wait margin whose zero level set is the release boundary. Monotone models, including min--max networks, partially monotone networks, calibrated lattices, and positive-derivative integral networks, provide established ways to encode order constraints \cite{sill1997monotonic,daniels2010monotone,gupta2016monotonic,you2017deep,wehenkel2019umnn}. We use this machinery only as a function class: the contribution is the combination of a Bellman-derived stopping margin with a conditional monotonicity constraint tied to physical urgency.


\section{Methodology}
\label{sec:method}

\begin{figure*}[t]
    \centering
    \includegraphics[width=\textwidth]{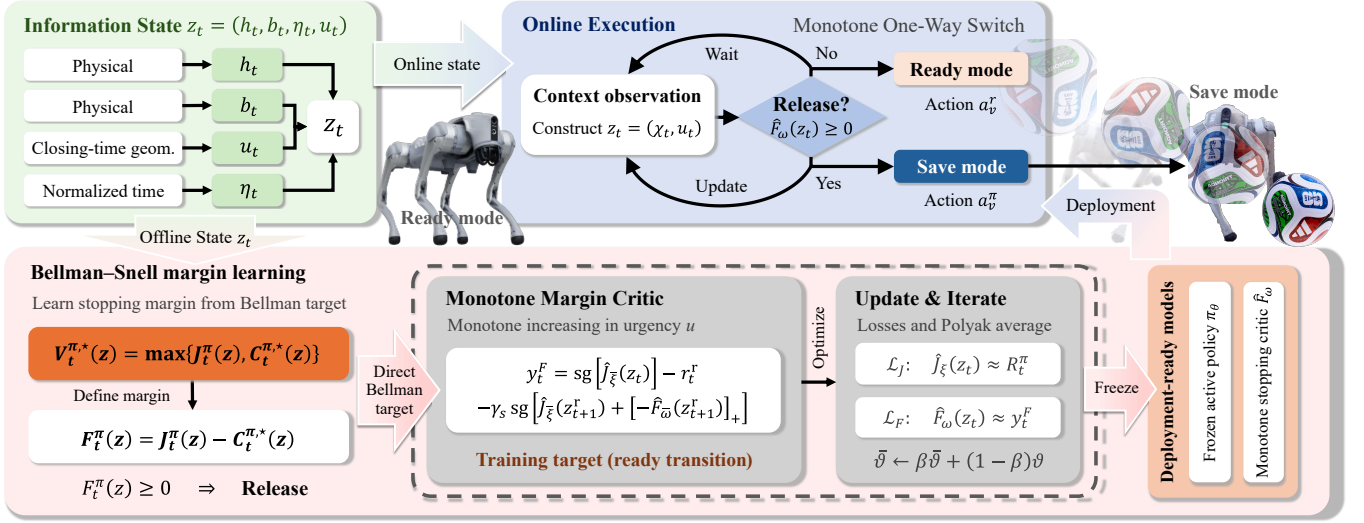}
    \caption{The MOS pipeline uses $z_t=(h_t,b_t,\eta_t,u_t)$ to combine robot history, target belief, decision time, and urgency; during training, active returns fit $\widehat J_\xi$ and ready transitions fit the Bellman stopping margin $\widehat F_\omega$ with monotonicity imposed only in $u_t$, while during deployment the robot remains ready for $\widehat F_\omega(z_t)<0$ and the first nonnegative margin releases the fixed closed-loop save policy.}
    \label{fig:method_overview}
\end{figure*}

Fig.~\ref{fig:method_overview} summarizes the two-stage method. The closed-loop save controller is trained with PPO and then held fixed; MOS learns only its release decision. A negative margin favors one more observation, whereas a nonnegative margin favors activating the save policy now.

\subsection{Problem Formulation}
\label{sec:problem}
Let $t\in\{0,\ldots,T\}$ index release decisions and let $G\in\mathcal G=\{1,\ldots,K\}$ denote the latent target region. From deployable history $\mathcal H_t$, a predictor produces the categorical belief
\begin{equation}
 b_t=\mathcal B(\mathcal H_t)\in\Delta^{K-1},\qquad b_t(g)\ge0,\quad \sum_{g\in\mathcal G}b_t(g)=1.
 \label{eq:belief}
\end{equation}

Let $h_t$ summarize recent robot and pre-contact observations, $u_t\in[0,1]$ denote physical urgency with larger values indicating less remaining opportunity, and $\eta_t=t/T$ denote normalized decision time. We define $\chi_t=(h_t,b_t,\eta_t)$ and the decision-scale information state
\begin{equation}
 z_t=(\chi_t,u_t)=(h_t,b_t,\eta_t,u_t).
 \label{eq:state}
\end{equation}

The predictor need only be deployable from $\mathcal H_t$; the formulation does not require an exactly calibrated Bayesian posterior. We assume $z_t$ is sufficient for the next ready-mode reward and transition and for the return of the fixed post-activation policy $\pi$. The policy $\pi$ is learned before MOS; its motor-level input, action, reward, and training settings are given in Sec.~\ref{sec:experiments}.

Let $\nu$ index the faster motor-control steps and $\iota(\nu)$ map motor step $\nu$ to its current release-decision index. With ready action $a_\nu^{\mathrm r}$, motor observation $x_\nu$, and one-way activation time $\sigma$, the hybrid controller is
\begin{equation}
 a_\nu=\begin{cases}
 a_\nu^{\mathrm r}, & \iota(\nu)<\sigma,\\
 a_\nu^\pi,\quad a_\nu^\pi\sim\pi(\cdot\mid x_\nu,z_{\iota(\nu)}), & \iota(\nu)\ge\sigma.
 \end{cases}
 \label{eq:hybrid_action}
\end{equation}

Thus activation changes the feedback mode rather than fixing an open-loop target: $\pi$ remains closed-loop and receives updated context after release. The stopping decision selects when to act, not which target to follow. Let $r_k^{\mathrm r}$ be the decision-scale reward accrued while waiting, $r_k$ the common task reward after release, and $\gamma_{\mathrm s}\in(0,1)$ the stopping discount. The value of immediate release is
\begin{equation}
J_t^\pi(z)=\mathbb E_\pi\!\left[\sum_{k=t}^{T}\gamma_{\mathrm s}^{k-t}r_k\,\middle|\,z_t=z,\sigma=t\right].
\label{eq:J}
\end{equation}

Let $\mathfrak T_{t,T}$ be the stopping times in $\{t,\ldots,T\}$ adapted to deployable history. The timing problem is
\begin{equation}
V_t^{\pi,\star}(z)=\sup_{\sigma\in\mathfrak T_{t,T}}\mathbb E\!\left[
\sum_{k=t}^{\sigma-1}\gamma_{\mathrm s}^{k-t}r_k^{\mathrm r}
+\gamma_{\mathrm s}^{\sigma-t}J_\sigma^\pi(z_\sigma)\,\middle|\,z_t=z\right],
\label{eq:stopping_objective}
\end{equation}
with compulsory release at $T$. The optimization is therefore only over release time for a fixed post-activation controller.

\subsection{MOS: Direct Bellman Stopping Margin}
For $t<T$, let $(r_t^{\mathrm r},z_{t+1}^{\mathrm r})$ denote one ready-mode transition. The continuation value and the finite-horizon Snell recursion are
\begin{align}
C_t^{\pi,\star}(z)&=\mathbb E\!\left[r_t^{\mathrm r}+\gamma_{\mathrm s}V_{t+1}^{\pi,\star}(z_{t+1}^{\mathrm r})\mid z_t=z\right],\label{eq:C}\\
V_t^{\pi,\star}(z)&=\max\{J_t^\pi(z),C_t^{\pi,\star}(z)\},\qquad V_T^{\pi,\star}=J_T^\pi.\label{eq:snell}
\end{align}

Define the stopping margin
\begin{equation}
F_t^\pi(z)=J_t^\pi(z)-C_t^{\pi,\star}(z).
\label{eq:F}
\end{equation}

Release is optimal when $F_t^\pi(z)\ge0$. Since
$V_{t+1}^{\pi,\star}=J_{t+1}^\pi+\left[-F_{t+1}^\pi\right]_+$, define the continuation option
$O_t^\pi=V_t^{\pi,\star}-J_t^\pi=\left[-F_t^\pi\right]_+$ for $t<T$ and $O_T^\pi=0$. Eliminating
$V^{\pi,\star}$ from Eq.~\eqref{eq:snell} then gives
\begin{align}
F_t^\pi(z)={}&J_t^\pi(z)
-\mathbb E\!\left[r_t^{\mathrm r}+\gamma_{\mathrm s}\left(
J_{t+1}^\pi(z_{t+1}^{\mathrm r})\right.\right.\nonumber\\
&\left.\left.\qquad\qquad +O_{t+1}^\pi(z_{t+1}^{\mathrm r})\right)
\,\middle|\, z_t=z\right].
\label{eq:direct_margin}
\end{align}
Equation~\eqref{eq:direct_margin} is algebraically equivalent to the Snell recursion but exposes the boundary quantity itself as the learned object, avoiding subtraction of two independently fitted stopping values at deployment.

\begin{proposition}[Policy-conditional stopping optimality]
\label{prop:snell}
Under the information-state sufficiency condition, Eqs.~\eqref{eq:C}--\eqref{eq:snell} solve Eq.~\eqref{eq:stopping_objective} for the fixed policy $\pi$, and an optimal release time is
\begin{equation}
\sigma_\pi^\star=\min\bigl(\{t<T:F_t^\pi(z_t)\ge0\}\cup\{T\}\bigr).
\label{eq:optimal_release}
\end{equation}
\end{proposition}

\subsection{MOS Learning and Deployment}
\label{sec:learning}
We impose the physical prior only along urgency: for fixed $\chi=(h,b,\eta)$, consuming more of one closing opportunity should not make waiting relatively more attractive. Choose knots $0=q_0<\cdots<q_M=1$. A network produces an offset $c_\omega(\chi)$ and positive interval slopes
\begin{equation}
d_{\omega,j}(\chi)=d_{\min}+\operatorname{softplus}(s_\omega(\chi,q_j))>0,\qquad d_{\min}>0,
\label{eq:slope}
\end{equation}

For $u\in[q_j,q_{j+1}]$, the learned margin is
\begin{align}
\widehat F_\omega(\chi,u)={}&c_\omega(\chi)
+\sum_{i=0}^{j-1}(q_{i+1}-q_i)d_{\omega,i}(\chi)\nonumber\\
&+(u-q_j)d_{\omega,j}(\chi).
\label{eq:monotone_critic}
\end{align}
where $\operatorname{softplus}(x)=\log(1+e^x)$. Hence $\widehat F_\omega$ is continuous and strictly increasing in $u$ for fixed $\chi$, so its zero set cannot fragment along an urgency slice; its dependence on robot state, belief, and decision time remains unrestricted.

Let $\widehat J_\xi$ estimate the immediate-release value and let $(\bar\xi,\bar\omega)$ be delayed parameters. A sampled ready transition yields the semi-gradient target
\begin{align}
y_t^F={}&\operatorname{sg}[\widehat J_{\bar\xi}(z_t)]-r_t^{\mathrm r}\nonumber\\
&-\gamma_{\mathrm s}\operatorname{sg}\!\left[\widehat J_{\bar\xi}(z_{t+1}^{\mathrm r})
+\left[-\widehat F_{\bar\omega}(z_{t+1}^{\mathrm r})\right]_+\right].
\label{eq:target}
\end{align}
where $\operatorname{sg}$ stops gradients and the option term is zero at $T$. The critic minimizes
\begin{equation}
\mathcal L_F(\omega)=\mathbb E\!\left[\ell_\delta(\widehat F_\omega(z_t)-y_t^F)\right],
\label{eq:loss}
\end{equation}
with Huber loss $\ell_\delta$; no binary release label is required. For a candidate ready state $z_t$, a counterfactual rollout that releases the frozen policy immediately yields $R_t^\pi=\sum_{k=t}^{T}\gamma_{\mathrm s}^{k-t}r_k$ and trains $\widehat J_\xi$ through $\mathcal L_J(\xi)=\mathbb E[\ell_\delta(\widehat J_\xi(z_t)-R_t^\pi)]$. The actor $\pi_\theta$ is first trained by PPO \cite{schulman2017ppo} and then frozen during timing learning. Delayed parameters are updated by $\bar\vartheta\leftarrow\beta\bar\vartheta+(1-\beta)\vartheta$ for $\vartheta\in\{\xi,\omega\}$. Online execution stays in ready mode while $\widehat F_\omega(z_t)<0$ and permanently activates $\pi_\theta$ at the first nonnegative margin or at $T$.

\subsection{Threshold Structure and Boundary Error}
\label{sec:theory}
For fixed $t$ and $\chi$, let $\mathcal I_t(\chi)=[u_t^-(\chi),u_t^+(\chi)]$ be an admissible urgency interval on which the counterfactual exact margin $F_t^\pi(\chi,u)$ is defined. The structural claim is restricted to such fixed-information slices.

\begin{assumption}[Strong single crossing]
\label{ass:crossing}
For fixed $\pi$, $t<T$, and $\chi$, the exact margin is continuous on $\mathcal I_t(\chi)$ and there exists $m_t(\chi)>0$ such that, for $u_1<u_2$,
\begin{equation}
F_t^\pi(\chi,u_2)-F_t^\pi(\chi,u_1)\ge m_t(\chi)(u_2-u_1).
\label{eq:strong_crossing}
\end{equation}
\end{assumption}

\begin{theorem}[Upper activation set]
\label{thm:upper}
Under Assumption~\ref{ass:crossing}, $\mathcal S_t^\pi(\chi)=\{u\in\mathcal I_t(\chi):F_t^\pi(\chi,u)\ge0\}$ is an upper set. If the endpoint margins straddle zero, the activation boundary $\varphi_t^\pi(\chi)=\inf\mathcal S_t^\pi(\chi)$ is the unique interior root.
\end{theorem}

The learned critic in Eq.~\eqref{eq:monotone_critic} has this ordered geometry by construction even when the physical prior is misspecified; in that case monotonicity is approximation bias rather than an optimality guarantee. To connect critic quality to boundary quality, define $\varepsilon_{F,t}=\|\widehat F_t-F_t^\pi\|_\infty$, $\varepsilon_{J,t}=\|\widehat J_t-J_t^\pi\|_\infty$, $\widehat O_t=[-\widehat F_t]_+$, $\varepsilon_{O,t}=\|\widehat O_t-O_t^\pi\|_\infty$, and the Bellman fitting residual $\delta_{B,t}=\|\widehat F_t-\widehat{\mathcal B}_t\|_\infty$, where $\widehat{\mathcal B}_t$ is the approximate version of Eq.~\eqref{eq:direct_margin}.

\begin{theorem}[Approximation-to-boundary error]
\label{thm:error}
With a shared ready-mode reward and transition law,
\begin{equation}
\varepsilon_{F,t}\le\delta_{B,t}+\varepsilon_{J,t}+\gamma_{\mathrm s}\varepsilon_{J,t+1}+\gamma_{\mathrm s}\varepsilon_{O,t+1},
\label{eq:error_recursion}
\end{equation}
where $\varepsilon_{O,t+1}\le\varepsilon_{F,t+1}$. Under uniform bounds $\delta_{B,t}\le\bar\delta_B$ and $\varepsilon_{J,t}\le\bar\varepsilon_J$,
\begin{equation}
\varepsilon_{F,t}\le[\bar\delta_B+(1+\gamma_{\mathrm s})\bar\varepsilon_J]\frac{1-\gamma_{\mathrm s}^{T-t}}{1-\gamma_{\mathrm s}}.
\label{eq:error_geometric}
\end{equation}

\begin{algorithm}[t]
\caption{Training and deployment of monotone optimal stopping (MOS) first train and freeze the save policy, then use active returns to fit $\widehat J_\xi$ and ready transitions to fit $\widehat F_\omega$, with the first nonnegative margin triggering release.}
\label{alg:method}
\begin{algorithmic}[1]
\REQUIRE Horizon $T$, knots $\{q_j\}_{j=0}^{M}$, update count $N_F$, averaging factor $\beta$
\STATE Train the active policy $\pi_\theta$ with exploratory release times
\STATE Fit $\widehat J_\xi$ to sampled active returns and freeze $\theta$
\FOR{each timing iteration}
    \STATE Collect ready transitions and counterfactual active returns
    \STATE Update $\xi$ from active-return regression
    \FOR{$j=1,\ldots,N_F$}
        \STATE Form $y_t^F$ by Eq.~\eqref{eq:target} and update $\omega$ using Eq.~\eqref{eq:loss}
    \ENDFOR
    \STATE Update delayed parameters by Polyak averaging with factor $\beta$
\ENDFOR
\FOR{$t=0,\ldots,T$ during deployment}
    \STATE Construct $z_t$ from deployable observations
    \IF{$\widehat F_\omega(z_t)\ge0$ or $t=T$}
        \STATE Permanently activate $\pi_\theta$ and disable the timing module
    \ENDIF
\ENDFOR
\end{algorithmic}
\end{algorithm}

\begin{table}[t]
\centering
\caption{The experimental setup and evaluation protocols summarize the platform, controller, and learning configurations.}
\label{tab:setup}
\tabfont
\renewcommand{\arraystretch}{1.04}
\setlength{\tabcolsep}{2.0pt}
\begin{tabular*}{\columnwidth}{@{\extracolsep{\fill}}llll@{}}
\toprule
\multicolumn{2}{c}{Category} &
\multicolumn{2}{c}{Configuration} \\
\cmidrule(r){1-2}\cmidrule(l){3-4}
Group & Item & Setting & Scale \\
\midrule
Protocol & P & Physics probe & $11{,}151$ eps.; 1 seed \\
Protocol & A & Timing benchmark & $2{,}000$/seed; 3 seeds \\
Protocol & C & Chase evaluation & $601$ shots; 1 policy \\
Robot & Platform & Go2 / Isaac Lab & 12 DoF \\
Control & Rates & Motor / timer & $50/10$ Hz \\
Policy & Input & $x_\nu,z_{\iota(\nu)}$ & GRU $128$ \\
Policy & Action & Joint targets & 12-D, $0.25$ scale \\
Policy & Reward & Intercept / stability & $+6/-5$ save/fall \\
Learning & Actor & $512$--$256$--$128$ & PPO \\
Learning & MOS heads & $256$--$256$ & $M=32$ \\
Learning & MOS opt. & Adam $3\!\times\!10^{-4}$ & $N_F=4$, $\beta=.995$ \\
\bottomrule
\end{tabular*}
\end{table}


\begin{figure*}[t]
    \centering
    \includegraphics{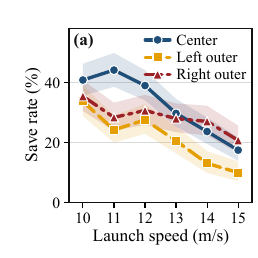}\hfill
    \includegraphics{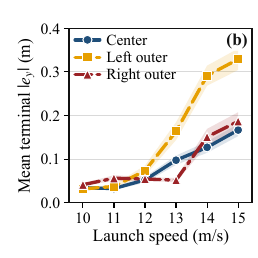}\hfill
    \includegraphics{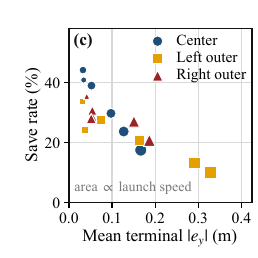}\hfill
    \includegraphics{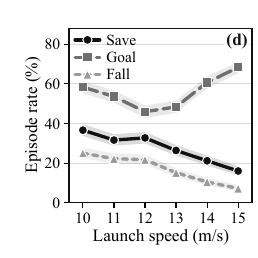}
    \caption{Protocol~P characterizes physical opportunity across launch speeds using (a) save rate by target band, (b) mean terminal lateral error $|e_y|$, (c) speed--band cells in the save--error plane, and (d) pooled save, goal, and fall rates, with bands and bars denoting $95\%$ intervals.}
    \label{fig:reachability}
\end{figure*}

\begin{table*}[t]
\centering
\caption{Protocol~A compares controlled timing methods in the upper block and changes relative to Policy-Gated together with gaps to the privileged Oracle in the lower block; C/S/E/Rev denote Central/Side/Extreme/Reversal.}
\label{tab:timing_benchmark}
\renewcommand{\arraystretch}{1.07}
\tabfont
\setlength{\tabcolsep}{2.45pt}
\begin{tabular*}{\textwidth}{@{\extracolsep{\fill}}ll rrrrr rrr}
\toprule
& & \multicolumn{5}{c}{Save / recovery (\%)} & \multicolumn{3}{c}{Aggregate (\%)}\\
\cmidrule(lr){3-7}\cmidrule(lr){8-10}
Method & Trigger & C & S & E & Rev & Rec.$^\dagger$ & Mean$\pm$SD & Lowest & Fall$\downarrow$\\
\midrule
Reactive & Contact & 72.1 & 30.6 & 4.8 & 32.4 & 31.2 & $35.0\pm1.8$ & 4.8 & 2.1\\
Fixed-Early & Fixed $-1.0$s & 86.8 & 74.6 & 58.9 & 29.4 & 22.6 & $62.4\pm1.5$ & 29.4 & 5.8\\
Confidence & Global conf. & 82.7 & 68.9 & 47.2 & 43.8 & 41.3 & $60.7\pm1.6$ & 43.8 & 4.2\\
Always-Active & Full window & 85.0 & 75.5 & 57.1 & 48.7 & 46.5 & $66.6\pm1.7$ & 48.7 & 6.0\\
Policy-Gated & Learned gate & 85.8 & 77.2 & 55.8 & 52.1 & 53.6 & $67.7\pm1.3$ & 52.1 & 3.5\\
Ours & Mono. margin & 87.4 & 80.8 & 63.0 & 66.5 & 64.7 & $74.4\pm1.1$ & 63.0 & 2.6\\
\midrule
Oracle$^\ddagger$ & True target & 89.7 & 82.9 & 67.6 & 86.0 & -- & 81.6 & 67.6 & 2.4\\
\bottomrule
\end{tabular*}

\vspace{0.65ex}
\tabfont
\setlength{\tabcolsep}{1.45pt}
\begin{tabular*}{\textwidth}{@{\extracolsep{\fill}}l rrrrrrr rrrrrr}
\toprule
& \multicolumn{7}{c}{Gain vs. Policy-Gated (pp)} & \multicolumn{6}{c}{Gap to privileged ref. (pp)}\\
\cmidrule(lr){2-8}\cmidrule(lr){9-14}
Method & $\Delta$C & $\Delta$S & $\Delta$E & $\Delta$Rev & $\Delta$Mean & $\Delta$Low & $\Delta$Fall & C & S & E & Rev & Mean & Low\\
\midrule
Reactive & -13.7 & -46.6 & -51.0 & -19.7 & -32.7 & -47.3 & +1.4 & 17.6 & 52.3 & 62.8 & 53.6 & 46.6 & 62.8\\
Fixed-Early & +1.0 & -2.6 & +3.1 & -22.7 & -5.3 & -22.7 & -2.3 & 2.9 & 8.3 & 8.7 & 56.6 & 19.2 & 38.2\\
Confidence & -3.1 & -8.3 & -8.6 & -8.3 & -7.0 & -8.3 & -0.7 & 7.0 & 14.0 & 20.4 & 42.2 & 20.9 & 23.8\\
Always-Active & -0.8 & -1.7 & +1.3 & -3.4 & -1.1 & -3.4 & -2.5 & 4.7 & 7.4 & 10.5 & 37.3 & 15.0 & 18.9\\
Policy-Gated & 0.0 & 0.0 & 0.0 & 0.0 & 0.0 & 0.0 & 0.0 & 3.9 & 5.7 & 11.8 & 33.9 & 13.9 & 15.5\\
Ours & +1.6 & +3.6 & +7.2 & +14.4 & +6.7 & +10.9 & +0.9 & 2.3 & 2.1 & 4.6 & 19.5 & 7.2 & 4.6\\
Oracle & +3.9 & +5.7 & +11.8 & +33.9 & +13.9 & +15.5 & +1.1 & 0.0 & 0.0 & 0.0 & 0.0 & 0.0 & 0.0\\
\midrule
Empirical gap reduction (\%) & 41.0 & 63.2 & 61.0 & 42.5 & 48.2 & 70.3 & 81.8 & \multicolumn{6}{c}{--}\\
\bottomrule
\end{tabular*}
\par\vspace{0.2ex}
\end{table*}

If Assumption~\ref{ass:crossing} holds, the exact boundary is interior, and
$\varepsilon_{F,t}<\min\{-F_t^\pi(\chi,u_t^-),F_t^\pi(\chi,u_t^+)\}$,
then the monotone approximate critic has a unique interior root $\widehat\varphi_t(\chi)$ and
\begin{equation}
|\widehat\varphi_t(\chi)-\varphi_t^\pi(\chi)|\le\frac{\varepsilon_{F,t}}{m_t(\chi)}.
\label{eq:root_error}
\end{equation}
\end{theorem}
\noindent Proof sketch.
Proposition~\ref{prop:snell} follows by backward induction on the finite horizon: each decision either releases with value $J_t^\pi$ or waits one ready-mode step and continues with $V_{t+1}^{\pi,\star}$. Theorem~\ref{thm:upper} follows from strict increase of $F_t^\pi(\chi,\cdot)$ under Assumption~\ref{ass:crossing}. For Theorem~\ref{thm:error}, add and subtract the approximate Bellman backup; the map $x\mapsto[-x]_+$ is $1$-Lipschitz, yielding Eq.~\eqref{eq:error_recursion}, whose backward unrolling gives Eq.~\eqref{eq:error_geometric}. Strong single crossing then converts margin error into the root bound in Eq.~\eqref{eq:root_error}. This is a fixed-slice boundary statement, not a bound on realized stopping-time differences along trajectories with evolving $\chi_t$.

\section{Results and Discussion}
\label{sec:results}

\subsection{Experimental Setup}
\label{sec:experiments}

Table~\ref{tab:setup} summarizes the shared platform, training configuration, and the three evaluation protocols used below. The active policy is trained with PPO before MOS. At motor step $\nu$, it receives fast robot feedback $x_\nu$ and context $z_{\iota(\nu)}=(h,b,\eta,u)$, and maps its 12-D action to joint targets as $q_\nu^\star=q_{\rm default}+0.25a_\nu^\pi$. Its $50$~Hz reward combines lateral/interception and pre-position objectives with posture, contact, joint-limit, and fall penalties, with alive $0.5$ per second, save $+6$, and fall $-5$. The actor/critic uses a $(512,256,128)$ ELU MLP with a 128-unit GRU; PPO uses discount $0.99$, GAE $0.95$, and clip $0.2$. Training randomizes friction $\mu\in[0.5,1.2]$, restitution $[0,0.15]$, trunk mass $[-1,3]$~kg, motor strength/gains, observation noise/latency, yaw $\pm10^\circ$, belief perturbations, and lateral pushes. The actor is frozen before MOS fitting, with the MOS optimizer settings summarized in the same setup.

Protocol~A uses $\{L_\ell,C_\ell,R_\ell,L_h,C_h,R_h\}$ regions and reports Central, Side, and Extreme aggregates. To isolate timing from perception, its controlled belief is
\begin{equation}
\lambda_t=\kappa(u_t)\mathbf e_G+s_b(u_t)\boldsymbol\varepsilon_t,
\qquad b_t=\operatorname{softmax}(\lambda_t/T_b),
\label{eq:belief_model}
\end{equation}
where $\kappa(u)=0.5+4u$, $s_b(u)=1.5-0.9u$, $T_b=1$, and $\boldsymbol\varepsilon_t$ is a zero-mean temporally correlated Gaussian process with correlation $0.8$. The simulator-only target $G$ is hidden from deployable timers and the shared motor policy; in reversal episodes the generator follows an incorrect target until $u_t=0.55$ and the true target thereafter.

\begin{figure*}[t]
    \centering
    \includegraphics{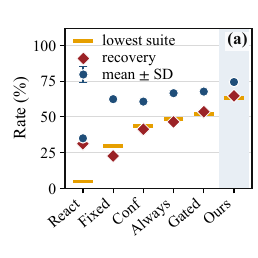}\hfill
    \includegraphics{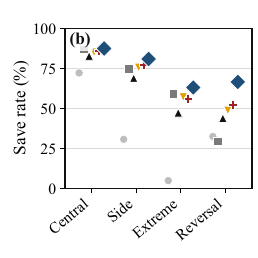}\hfill
    \includegraphics{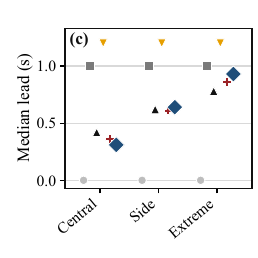}\hfill
    \includegraphics{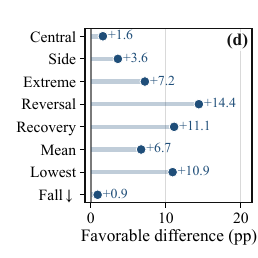}\\[0.2ex]
    \includegraphics{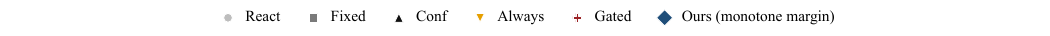}
    \caption{Protocol~A compares release rules using (a) mean save rate, lowest-suite save rate, and recovery, (b) save rate across Central, Side, Extreme, and Reversal suites, (c) median pre-contact release lead, and (d) the difference between MOS and Policy-Gated, where rightward values favor MOS; only the overall mean reports cross-seed SD and recovery is descriptive.}
    \label{fig:benchmark}
\end{figure*}

\begin{figure}[t]
    \centering
    \includegraphics[width=\columnwidth]{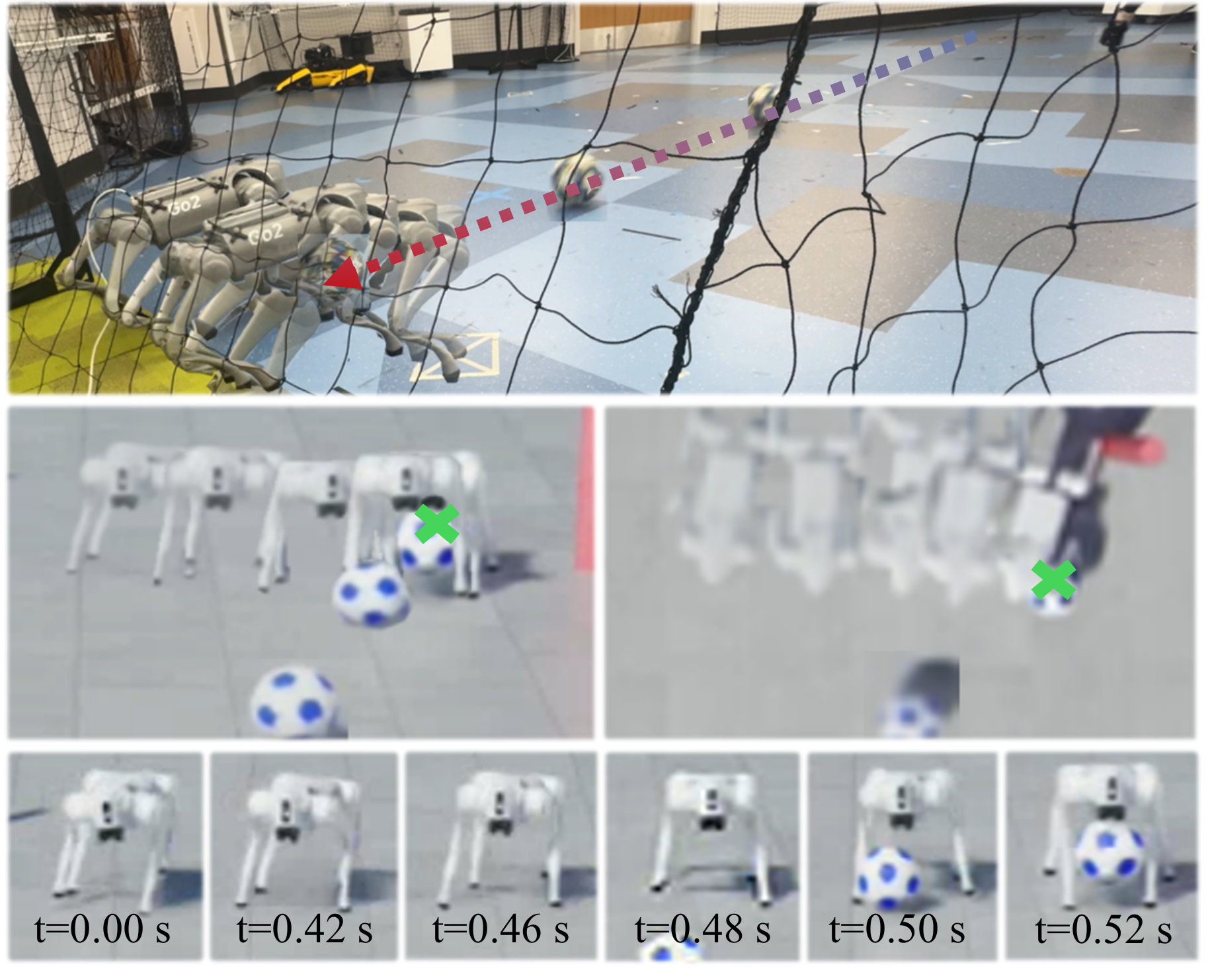}
    \caption{The hardware lateral-interception sequence shows an overlaid trajectory and close view of the save in the top and middle panels and front-view frames from $t=0$ to $0.52$~s in the bottom row, illustrating the rapid motion generated by the active policy.}
    \label{fig:hardware_fast}
\end{figure}

\begin{figure}[t]
    \centering
    \includegraphics{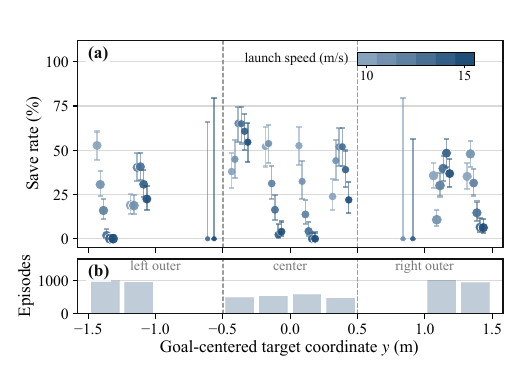}
    \caption{Protocol~P resolves save rate over goal-centered target position, with marker area indicating episode count, vertical bars showing Wilson $95\%$ intervals, and the lower strip showing sampling density; unsampled cells remain blank.}
    \label{fig:position}
\end{figure}

Urgency is the normalized bounded closing-time proxy
\begin{equation}
\begin{aligned}
\widetilde\tau_t&=\frac{d_t-d_c}{\max(-\dot d_t,v_{\min})},\qquad
\widehat\tau_t=\operatorname{clip}(\widetilde\tau_t,0,\tau_{\max}),\\
u_t&=1-\widehat\tau_t/\tau_{\max}.
\end{aligned}
\label{eq:urgency}
\end{equation}
with $\tau_{\max}=1.2$~s; $d_t$, $d_c$, and $\dot d_t$ are foot--ball separation, contact separation, and separation rate. Protocols~P/C diagnose controller physics and physical opportunity, while Protocol~A evaluates the timer.

\subsection{Opportunity Geometry and Timing Benchmark}
Protocol~P characterizes the loss of physical opportunity with shot difficulty. Fig.~\ref{fig:reachability} shows save rate falling from $36.6\%$ at $10$~m/s to $16.0\%$ at $15$~m/s while mean terminal $|e_y|$ rises from $0.036$ to $0.226$~m. At $15$~m/s the goal rate exceeds $68\%$ but falls remain $7\%$, indicating missed interception rather than instability. Table~\ref{tab:timing_benchmark} and Fig.~\ref{fig:benchmark} compare release rules. MOS reaches $74.4\pm1.1\%$ mean save, $+6.7$~pp over Policy-Gated, and improves Reversal from $52.1\%$ to $66.5\%$. Reversal is especially diagnostic because the early cue is intentionally wrong before switching to the true target; success therefore requires early motion without locking onto that cue. Its median release lead increases from $0.31$~s on Central to $0.93$~s on Extreme, supporting state-dependent rather than uniformly early activation.

Fig.~\ref{fig:hardware_fast} provides a real-robot example of the rapid lateral motion produced by the active policy and motivates preserving pre-contact opportunity. Fig.~\ref{fig:position} further shows substantial difficulty variation within each target band, supporting state- and urgency-conditioned timing rather than a single global confidence threshold. Protocol~P characterizes physical opportunity, while Protocol~A directly evaluates the release mechanism.

\subsection{Robustness and Closed-Loop Adaptation}
\begin{figure*}[tbp]
    \centering
    \includegraphics{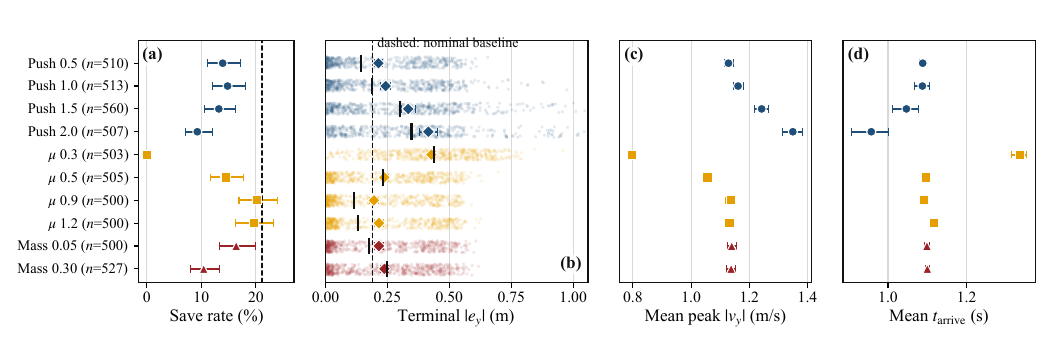}
    \caption{Protocol~P stress diagnostics at $14$~m/s report (a) save rate, (b) terminal-error distribution, (c) mean peak lateral speed, and (d) mean arrival time, with $95\%$ intervals and dashed references indicating the nominal condition.}
    \label{fig:diagnostics}
\end{figure*}

\begin{figure}[htbp]
    \centering
    \includegraphics{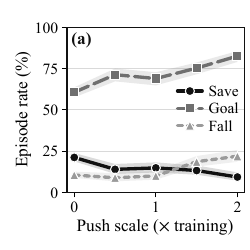}\hfill
    \includegraphics{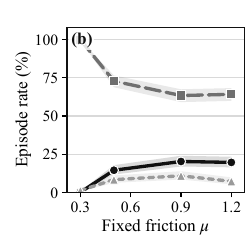}\\[0.4ex]
    \includegraphics{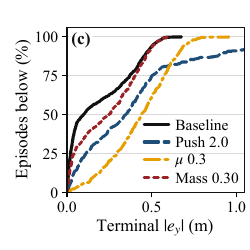}\hfill
    \includegraphics{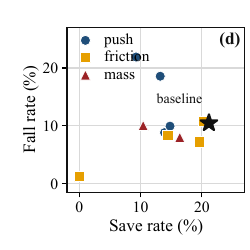}
    \caption{Protocol~P stress sweeps at $14$~m/s evaluate (a) lateral pushes, (b) fixed friction, (c) terminal-error cumulative distributions, and (d) fall rate versus save rate, showing that pushes mainly destabilize recovery whereas very low friction reduces lateral authority.}
    \label{fig:robustness}
\end{figure}

\begin{figure}[htbp]
    \centering
    \includegraphics[width=\columnwidth]{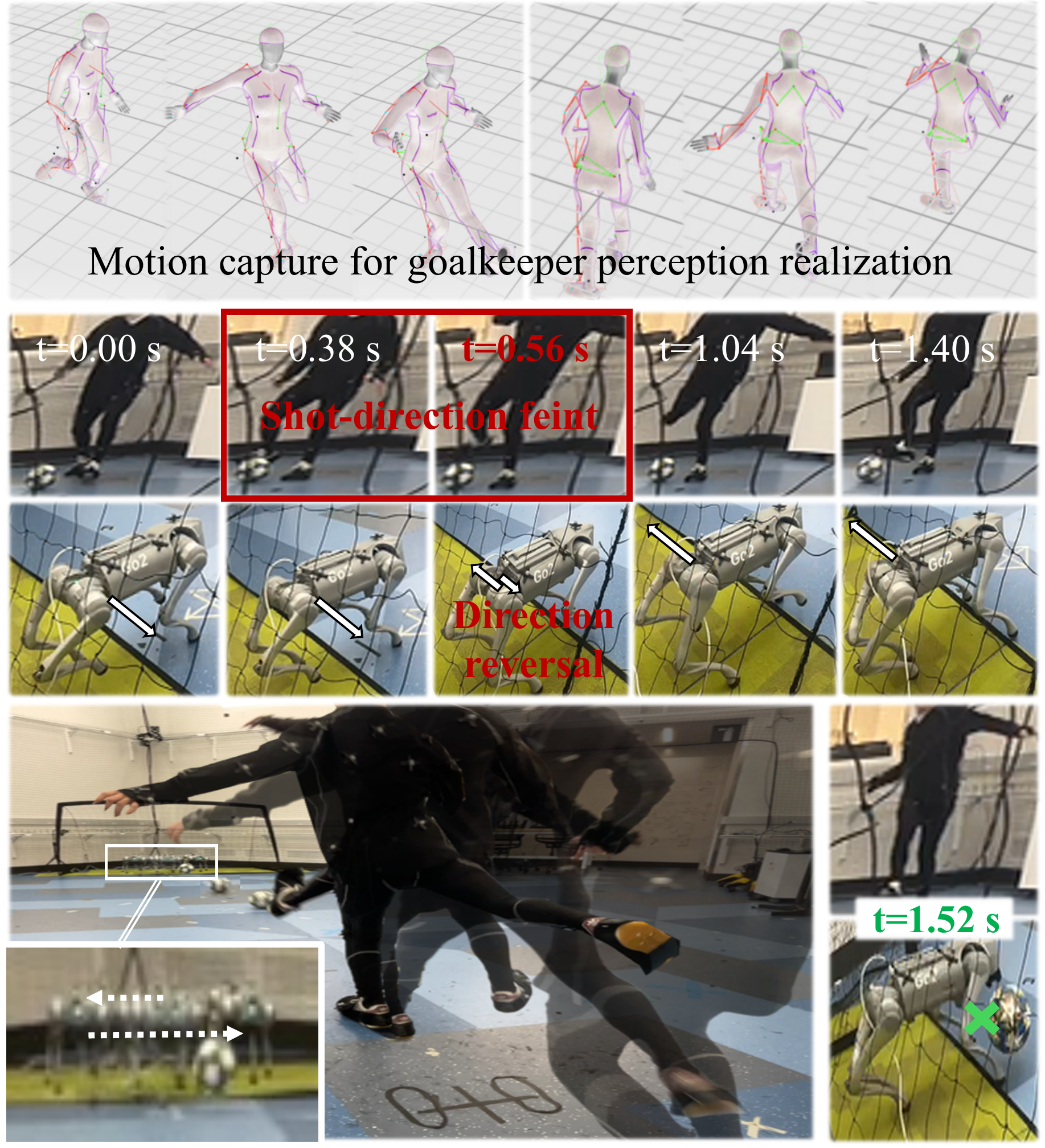}
    \caption{In the hardware human shot-direction feint, the keeper releases on an early cue, reverses after the cue changes, and reaches the ball at $t=1.52$~s, demonstrating that release activates a closed-loop save policy rather than committing to a fixed target.}
    \label{fig:hardware_reversal}
\end{figure}

Fig.~\ref{fig:diagnostics} shows controller-level stress responses at $14$~m/s. A $2\times$ push reduces save rate from $21.2\%$ to $9.3\%$ and raises falls from $10.5\%$ to $21.9\%$, whereas $\mu=0.3$ gives $0\%$ saves with only $1.2\%$ falls. Fig.~\ref{fig:robustness} further separates the failure modes, showing that pushes reduce saves while increasing falls whereas very low friction collapses saves with little increase in falls, indicating loss of lateral authority. Fig.~\ref{fig:hardware_reversal} verifies this closed-loop behavior under a human shot-direction feint: the keeper releases on an early cue, reverses after it changes, and reaches the ball.

\begin{table}[htbp]
\centering
\caption{Protocol~C reports closed-loop outcomes by aim band.}
\label{tab:chase}
\tabfont
\renewcommand{\arraystretch}{1.04}
\setlength{\tabcolsep}{2.0pt}
\begin{tabular*}{\columnwidth}{@{\extracolsep{\fill}}lccccc@{}}
\toprule
\multicolumn{2}{c}{Evaluation} & \multicolumn{4}{c}{Outcome (\%)} \\
\cmidrule(r){1-2}\cmidrule(l){3-6}
Aim band & $n$ & Contact & Save & On-target & Fall \\
\midrule
Left   & $207$ & $57.0$ & $32.4$ & $38.1$ & $13.0$ \\
Center & $187$ & $80.2$ & $40.1$ & $41.9$ & $25.7$ \\
Right  & $207$ & $70.5$ & $35.7$ & $41.6$ & $19.3$ \\
\midrule
All    & $601$ & $68.9$ & $35.9$ & $40.5$ & $19.1$ \\
\bottomrule
\end{tabular*}
\tabnote[\columnwidth]{Except for $n$, all entries are percentages. On-target is the save rate after excluding the 68 shots that missed the frame untouched.}
\end{table}

\begin{figure}[htbp]
    \centering
    \includegraphics[width=\columnwidth]{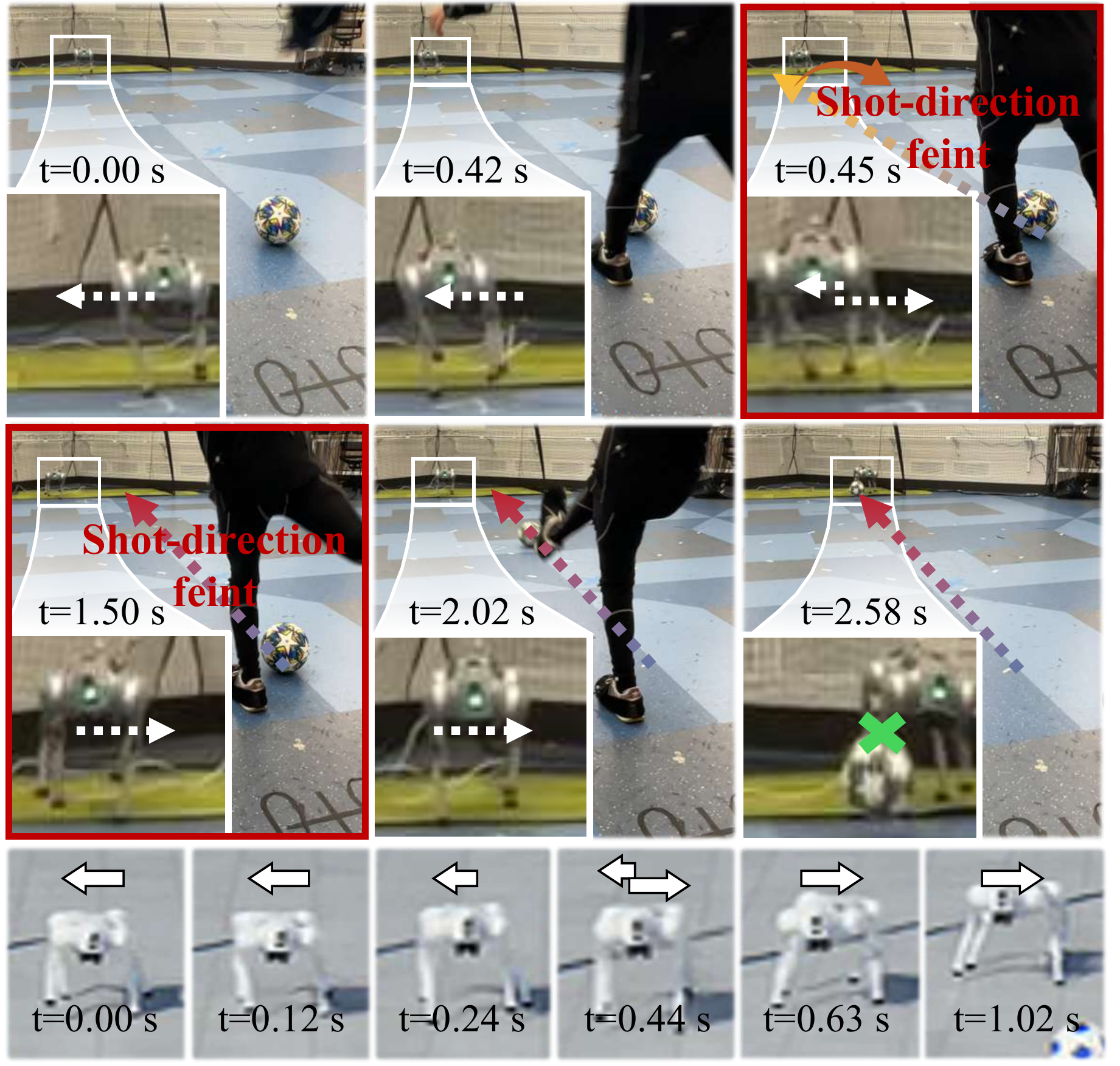}
    \caption{In the extended human-feint trial, the robot first moves left, reverses near $t=0.44$~s, and then moves right before the final save, demonstrating repeated post-release redirection after a single anticipatory release.}
    \label{fig:hardware_multifeint}
\end{figure}

\begin{figure}[htbp]
    \centering
    \includegraphics{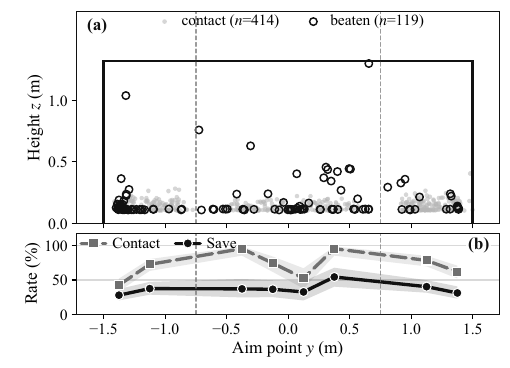}
    \caption{Protocol~C maps goal-mouth performance using (a) ball--keeper contacts and untouched on-target crossings and (b) contact and save rates versus lateral aim point.}
    \label{fig:goalmouth}
\end{figure}

Fig.~\ref{fig:hardware_multifeint} shows an extended human feint with repeated post-release redirection after a single release. Table~\ref{tab:chase} summarizes the corresponding Protocol~C outcomes by target band. Fig.~\ref{fig:goalmouth} localizes the residual failure. Contacts remain below $z=0.41$~m while untouched on-target crossings reach $1.30$~m; contact is $80.2\%$ center, $70.5\%$ right, and $57.0\%$ left. Protocol~A isolates timing under controlled beliefs, while Protocols~P/C characterize physical opportunity and controller failure modes. Hardware trials demonstrate anticipatory release and post-release redirection under human feints. Optimality remains conditional on the fixed post-activation policy and one-way architecture.

\section{Conclusion}
Anticipatory goalkeeping couples improving information with vanishing reachability, while human feints can overturn early cues. MOS resolves this conflict by comparing acting now with waiting one more observation, ordering this preference by urgency while preserving closed-loop adaptation after release.
\begin{itemize}
    \item A policy-conditional finite-horizon optimal-stopping formulation converts anticipatory release into an explicit act-versus-wait decision, and a direct Bellman recursion enables the corresponding stopping margin to be learned without binary timing labels.
    \item The proposed MOS critic imposes monotonicity only along physical urgency while retaining unrestricted dependence on robot state, target belief, and decision time. Under strong single crossing, this structure yields a threshold release boundary and connects value and Bellman approximation errors to release-boundary error.
    \item Integrated with a PPO-trained 12-DoF quadruped save policy, MOS increases mean save rate from $67.7\%$ to $74.4\%$ and matched reversal save rate from $52.1\%$ to $66.5\%$ over a parameter-matched learned gate. Hardware human-feint trials demonstrate post-release correction after misleading early cues, confirming that early activation preserves both interception time and closed-loop adaptability.

\end{itemize}


\balance
\bibliographystyle{IEEEtran}
\bibliography{references}

@article{kaelbling1998pomdp,
  title = {Planning and Acting in Partially Observable Stochastic Domains},
  author = {Kaelbling, Leslie Pack and Littman, Michael L. and Cassandra, Anthony R.},
  journal = {Artificial Intelligence},
  volume = {101},
  number = {1--2},
  pages = {99--134},
  year = {1998}
}

@article{lauri2023pomdp,
  title = {Partially Observable Markov Decision Processes in Robotics: A Survey},
  author = {Lauri, Mikko and Hsu, David and Pajarinen, Joni},
  journal = {IEEE Transactions on Robotics},
  volume = {39},
  number = {1},
  pages = {21--40},
  year = {2023},
  doi = {10.1109/TRO.2022.3200138}
}

@article{wang2017anticipatory,
  title = {Anticipatory Action Selection for Human--Robot Table Tennis},
  author = {Wang, Zhikun and Boularias, Abdeslam and M{\"u}lling, Katharina and Sch{\"o}lkopf, Bernhard and Peters, Jan},
  journal = {Artificial Intelligence},
  volume = {247},
  pages = {399--414},
  year = {2017}
}

@book{peskir2006optimal,
  title = {Optimal Stopping and Free-Boundary Problems},
  author = {Peskir, Goran and Shiryaev, Albert},
  publisher = {Birkh{\"a}user},
  address = {Basel},
  year = {2006}
}

@article{tsitsiklis2001regression,
  title = {Regression Methods for Pricing Complex American-Style Options},
  author = {Tsitsiklis, John N. and Van Roy, Benjamin},
  journal = {IEEE Transactions on Neural Networks},
  volume = {12},
  number = {4},
  pages = {694--703},
  year = {2001},
  doi = {10.1109/72.935083}
}

@inproceedings{sill1997monotonic,
  title = {Monotonic Networks},
  author = {Sill, Joseph},
  booktitle = {Advances in Neural Information Processing Systems},
  volume = {10},
  year = {1997}
}

@article{daniels2010monotone,
  title = {Monotone and Partially Monotone Neural Networks},
  author = {Daniels, Hennie and Velikova, Marina},
  journal = {IEEE Transactions on Neural Networks},
  volume = {21},
  number = {6},
  pages = {906--917},
  year = {2010},
  doi = {10.1109/TNN.2010.2044803}
}

@article{gupta2016monotonic,
  title = {Monotonic Calibrated Interpolated Look-Up Tables},
  author = {Gupta, Maya and Cotter, Andrew and Pfeifer, Jan and Voevodski, Konstantin and Canini, Kevin and Mangylov, Alexander and Moczydlowski, Wojciech and van Esbroeck, Alexander},
  journal = {Journal of Machine Learning Research},
  volume = {17},
  number = {109},
  pages = {1--47},
  year = {2016}
}

@inproceedings{you2017deep,
  title = {Deep Lattice Networks and Partial Monotonic Functions},
  author = {You, Seungil and Ding, David and Canini, Kevin and Pfeifer, Jan and Gupta, Maya},
  booktitle = {Advances in Neural Information Processing Systems},
  volume = {30},
  year = {2017}
}

@inproceedings{wehenkel2019umnn,
  title = {Unconstrained Monotonic Neural Networks},
  author = {Wehenkel, Antoine and Louppe, Gilles},
  booktitle = {Advances in Neural Information Processing Systems},
  volume = {32},
  year = {2019}
}

@article{becker2019deep,
  title = {Deep Optimal Stopping},
  author = {Becker, Sebastian and Cheridito, Patrick and Jentzen, Arnulf},
  journal = {Journal of Machine Learning Research},
  volume = {20},
  number = {74},
  pages = {1--25},
  year = {2019}
}

@inproceedings{chen2020stop,
  title = {Learning to Stop While Learning to Predict},
  author = {Chen, Xinshi and Dai, Hanjun and Li, Yu and Gao, Xin and Song, Le},
  booktitle = {Proceedings of the 37th International Conference on Machine Learning},
  series = {Proceedings of Machine Learning Research},
  volume = {119},
  pages = {1520--1530},
  year = {2020}
}

@inproceedings{huang2023goalkeeper,
  title = {Creating a Dynamic Quadrupedal Robotic Goalkeeper with Reinforcement Learning},
  author = {Huang, Xiaoyu and Li, Zhongyu and Xiang, Yanzhen and Ni, Yiming and Chi, Yufeng and Li, Yunhao and Yang, Lizhi and Peng, Xue Bin and Sreenath, Koushil},
  booktitle = {2023 IEEE/RSJ International Conference on Intelligent Robots and Systems},
  pages = {2715--2722},
  year = {2023}
}

@article{ren2025humanoidgoalkeeper,
  title = {Humanoid Goalkeeper: Learning from Position Conditioned Task-Motion Constraints},
  author = {Ren, Junli and Long, Junfeng and Huang, Tao and Wang, Huayi and Wang, Zirui and Jia, Feiyu and Zhang, Wentao and Wang, Jingbo and Luo, Ping and Pang, Jiangmiao},
  journal = {arXiv preprint arXiv:2510.18002},
  year = {2025}
}

@article{haarnoja2024soccer,
  title = {Learning Agile Soccer Skills for a Bipedal Robot with Deep Reinforcement Learning},
  author = {Haarnoja, Tuomas and Moran, Ben and Lever, Guy and Huang, Sandy H. and Tirumala, Dhruva and Humplik, Jan and Wulfmeier, Markus and Tunyasuvunakool, Saran and Siegel, Noah Y. and Hafner, Roland and Bloesch, Michael and Hartikainen, Kristian and Byravan, Arunkumar and Hasenclever, Leonard and Tassa, Yuval and Sadeghi, Fereshteh and Batchelor, Nathan and Casarini, Federico and Saliceti, Stefano and Game, Charles and Sreendra, Neil and Patel, Kushal and Gwira, Marlon and Huber, Andrea and Hurley, Nicole and Nori, Francesco and Hadsell, Raia and Heess, Nicolas},
  journal = {Science Robotics},
  volume = {9},
  number = {89},
  pages = {eadn1844},
  year = {2024}
}

@inproceedings{rudin2022walk,
  title = {Learning to Walk in Minutes Using Massively Parallel Deep Reinforcement Learning},
  author = {Rudin, Nikita and Hoeller, David and Reist, Philipp and Hutter, Marco},
  booktitle = {Proceedings of the 5th Conference on Robot Learning},
  series = {Proceedings of Machine Learning Research},
  volume = {164},
  pages = {91--100},
  year = {2022}
}

@article{schulman2017ppo,
  title = {Proximal Policy Optimization Algorithms},
  author = {Schulman, John and Wolski, Filip and Dhariwal, Prafulla and Radford, Alec and Klimov, Oleg},
  journal = {arXiv preprint arXiv:1707.06347},
  year = {2017}
}

@article{zhang2025multi,
  title={Multi-scale reinforcement learning of dynamic energy controller for connected electrified vehicles},
  author={Zhang, Hao and Lei, Nuo and Li, Shengbo Eben and Zhang, Junzhi and Wang, Zhi},
  journal={IEEE Transactions on Intelligent Transportation Systems},
  year={2025},
  publisher={IEEE}
}

@article{zhang2025bi,
  title={Bi-level transfer learning for lifelong-intelligent energy management of electric vehicles},
  author={Zhang, Hao and Lei, Nuo and Peng, Wang and Li, Bingbing and Lv, Shujun and Chen, Boli and Wang, Zhi},
  journal={IEEE Transactions on Intelligent Transportation Systems},
  volume={26},
  number={10},
  pages={16174--16187},
  year={2025},
  publisher={IEEE}
}

@article{zhang2026cognition,
  title={Cognition to Control-Multi-Agent Learning for Human-Humanoid Collaborative Transport},
  author={Zhang, Hao and Zhao, Ding and Tseng, H Eric},
  journal={arXiv preprint arXiv:2603.03768},
  year={2026}
}

@article{zhang2026interaction,
  title={Interaction-Aware Whole-Body Control for Compliant Object Transport},
  author={Zhang, Hao and Tseng, Yves and Zhao, Ding and Tseng, H Eric},
  journal={arXiv preprint arXiv:2603.03751},
  year={2026}
}

@article{zhang2026halo,
  title={HALO: Learning Human-Robot Collaboration via Heterogeneous-Agent Lyapunov Policy Optimization},
  author={Zhang, Hao and Niu, Yaru and Wang, Yikai and Zhao, Ding and Tseng, H Eric},
  journal={arXiv preprint arXiv:2603.03741},
  year={2026},
  publisher={International Conference on Machine Learning (ICML)}
}


\end{document}